\documentclass{bmvc2k}

\title{\hspace*{1.2cm}\parbox{\dimexpr\linewidth-1.2cm\relax}{%
  \Large \bfseries Image Classifiers are Efficient Self-Supervised Video Representation Learners%
}}

\addauthor{Owais Iqbal}{owais.iqbal@kgpian.iitkgp.ac.in}{1}
\addauthor{Sudipta Sarkar}{sudipta25t@kgpian.iitkgp.ac.in}{1}
\addauthor{Shyam Marjit}{shyammarjit@iisc.ac.in}{2}
\addauthor{Omprakash Chakraborty}{omprakash.chakraborty@livia.etsmtl.ca}{3}
\addauthor{Anirban Chakraborty}{anirban@iisc.ac.in}{2}
\addauthor{Abir Das}{abir@cse.iitkgp.ac.in}{1}

\addinstitution{
 Indian Institute of Technology\\
 Kharagpur, India
}

\addinstitution{
 Indian Institute of Science\\
 Bangalore, India
}

\addinstitution{
 \'Ecole de technologie sup\'erieure\\
 Montreal, Canada
}

\runninghead{Iqbal et al.}{VideoMSN}

\usepackage{graphicx}
\usepackage{booktabs}
\usepackage[accsupp]{axessibility}
\usepackage{amssymb}
\usepackage{multirow}
\def\ours{\texttt{\textbf{VideoMSN}}\xspace}

\makeatletter
\newcommand{\sublabelabl}[2]{%
  \Hy@raisedlink{\hypertarget{#1}{}}%
  \protected@write\@auxout{}{%
    \string\newlabel{#1}{%
      {\ref{tab:ablations} (#2)}%
      {\thepage}%
      {Table \ref{tab:ablations} (#2)}%
      {#1}%
      {}%
    }%
  }%
}
\makeatother

\usepackage{tikz}
\usepackage{graphicx}
\usepackage{float}
\usepackage{amsmath}
\usepackage{amssymb}
\usepackage{mathtools}
\usepackage{booktabs}
\usepackage{multirow}
\usepackage{makecell}
\usepackage{adjustbox}
\usepackage{tabularx}
\usepackage{pifont}
\usepackage[compatibility=false]{caption}
\usepackage{colortbl}
\usepackage{tcolorbox}
\usepackage[flushleft]{threeparttable}
\usepackage{xcolor}
\usepackage{wrapfig}
\usepackage[normalem]{ulem}
\usepackage{arydshln}

\makeatletter

\newcommand{\sublabel}[2]{%
  \Hy@raisedlink{\hypertarget{#1}{}}%
  \protected@write\@auxout{}{%
    \string\newlabel{#1}{%
      {\ref{tab:ablations} (#2)}%
      {\thepage}%
      {Table \ref{tab:ablations} (#2)}%
      {#1}%
      {}%
    }%
  }%
}

\newcommand{\sublabelnew}[2]{%
  \Hy@raisedlink{\hypertarget{#1}{}}%
  \protected@write\@auxout{}{%
    \string\newlabel{#1}{%
      {\ref{tab:new_ablations} (#2)}%
      {\thepage}%
      {Table \ref{tab:new_ablations} (#2)}%
      {#1}%
      {}%
    }%
  }%
}

\makeatother

\newcommand*\colourcheck[1]{%
  \expandafter\newcommand\csname #1check\endcsname{%
    \textcolor{#1}{\ding{52}}%
  }%
}

\colourcheck{blue}
\colourcheck{green}
\colourcheck{red}

\newlength{\savewidth}

\newcommand{\shline}{%
  \noalign{%
    \global\savewidth\arrayrulewidth
    \global\arrayrulewidth 1pt%
  }%
  \hline
  \noalign{%
    \global\arrayrulewidth\savewidth
  }%
}

\newcolumntype{x}[1]{%
  >{\centering\arraybackslash}p{#1pt}%
}

\newcolumntype{y}[1]{%
  >{\raggedright\arraybackslash}p{#1pt}%
}

\newcolumntype{z}[1]{%
  >{\raggedleft\arraybackslash}p{#1pt}%
}

\newcommand{\xmark}{\ding{55}}%

\definecolor{citecolor}{RGB}{34,139,34}
\definecolor{citecolor2}{HTML}{0071bc}
\definecolor{lightred}{RGB}{241,140,142}
\definecolor{carmine}{rgb}{0.59,0.0,0.09}
\definecolor{grey}{rgb}{0.6,0.6,0.6}

\newcommand{\pacc}[1]{%
  {\bfseries
  \fontsize{7.5}{42}\selectfont
  \color{citecolor!80}~(#1)}%
}

\newcommand{\macc}[1]{%
  {\bfseries
  \fontsize{7.5}{42}\selectfont
  \color{lightred!180}~(#1)}%
}

\makeatletter

\def\adl@drawiv#1#2#3{%
  \hskip.5\tabcolsep
  \xleaders#3{%
    #2.5\@tempdimb #1{1}#2.5\@tempdimb%
  }%
  #2\z@ plus1fil minus1fil\relax
  \hskip.5\tabcolsep
}

\newcommand{\cdashlinelr}[1]{%
  \noalign{%
    \vskip\aboverulesep
    \global\let\@dashdrawstore\adl@draw
    \global\let\adl@draw\adl@drawiv
  }%
  \cdashline{#1}%
  \noalign{%
    \global\let\adl@draw\@dashdrawstore
    \vskip\belowrulesep
  }%
}

\makeatother

\usepackage{titletoc}

\usepackage{color}
\usepackage{orcidlink}
\usepackage[title]{appendix}

\begin{document}
{
\setlength{\leftskip}{1.2cm}
\maketitle
}

\begin{abstract}
We introduce \ours, a Masked Siamese Network framework for efficient self-supervised spatio-temporal representation learning in videos.
Instead of relying on heavy 3D architectures or reconstruction-based autoencoders for learning with unlabeled data, we repurpose standard image Vision Transformers by representing videos as super images which are grids composed of frames sampled from videos. From each super image, we construct two views: one with spatial patch masking and the other with temporal frame masking, ensuring no information leakage across frames. A shared Vision Transformer (ViT) encoder aligns their embeddings using a masked Siamese loss, capturing both motion and appearance cues without reconstruction. 
Our decoder-free formulation leverages an image foundation model towards efficient video representation learning.
Starting from pretrained DINO-v3 and DeiT-v3 image encoders, \ours achieves state-of-the-art performance on Kinetics-400, UCF101, and HMDB51 while requiring up to \textbf{$32\times$ fewer} and \textbf{$160\times$ fewer} video pretraining epochs compared to prior video self-supervised learning methods.
Our proposed approach also shows strong performance in
low-shot classification, confirming the transferability of the learned representations in a label-scarce scenario.
Project Page: \href{https://cvir.github.io/projects/videomsn}{https://cvir.github.io/projects/videomsn}.

\end{abstract}
\section{Introduction}
\label{sec:intro}
Self-supervised learning (SSL) has emerged as a strong paradigm for visual representation learning without the need for meticulously labeled data.
Instead, it intelligently leverages patterns naturally present in images and learns through pretext tasks.
Pretext tasks for images can vary from exploiting their spatial structure~\cite{doersch2015unsupervised, noroozi2017representation, dosovitskiy2016discriminative}, solving jigsaw puzzles~\cite{noroozi2016unsupervised}, colorizing images~\cite{zhang2016colorful, larsson2017colorization}, predicting rotations in artificially rotated images~\cite{gidaris2018unsupervised}, \textit{etc}.
One of the common and effective pretext tasks is Masked Image Modeling (MIM).
It involves masking a portion of the input and either predicting the masked regions~\cite{he2022masked, pmlr-v235-bar24a, assran2023self} or producing similar embeddings for the masked and the unmasked inputs~\cite{assran2022masked, tao2023siamese}.
While reconstructing masked patches through an encoder-decoder framework is effective, this approach emphasizes unnecessary low-level details, often at the expense of longer training and higher compute costs.
In contrast, a Masked Siamese Network (MSN)~\cite{assran2022masked} uses an encoder-only framework to align features of masked and unmasked views of the same image.
As masked image portions do not go through the encoder, it exhibits good computational scaling with fewer training epochs compared to the reconstruction-based approaches.

\begin{wrapfigure}{r}{0.55\textwidth}
\vspace{-8pt}
\centering
\includegraphics[width=\linewidth]{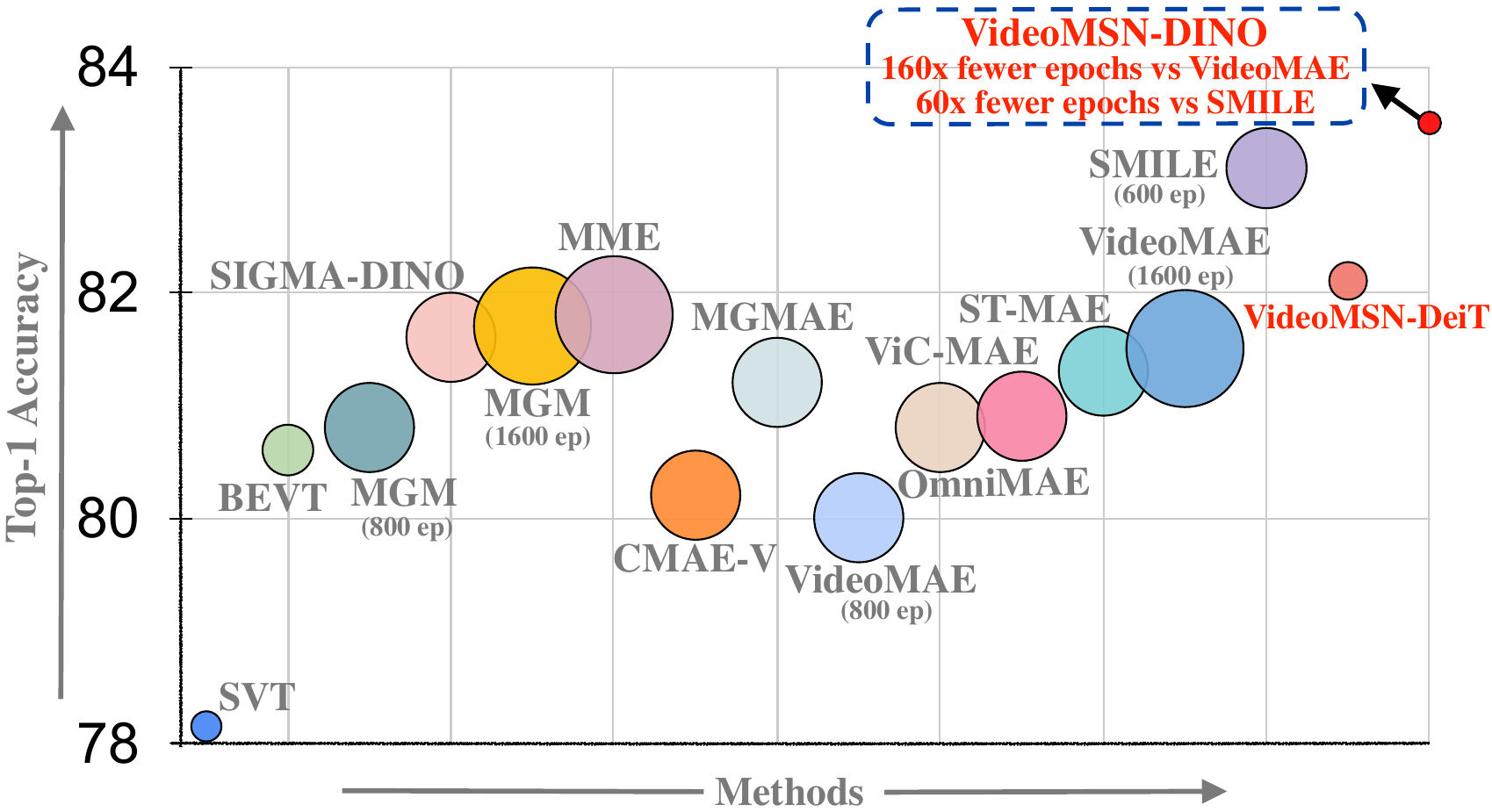}
\vspace{-7pt}
\caption{Comparison of top-1 accuracy on Kinetics-400 across state-of-the-art self-supervised video representation learning methods. Each point denotes a method, with bubble size proportional to its pretraining epochs. Our method, \ours with DINO-v3 backbone, achieves state-of-the-art performance with \textbf{$\textbf{160}\times$} and $\textbf{60}\times$ fewer training epochs compared to VideoMAE~\cite{tong2022videomae} and SMILE~\cite{thoker2025smile} respectively.
}

\vspace{-6pt}
\label{fig:teaser}
\end{wrapfigure}

Yet efforts to scale these methods to videos are hindered by
the need to capture both spatial and temporal dynamics.
Spatio-temporal processing often relies on heavy 3D CNNs or video vision transformers~\cite{arnab2021vivit, bertasius2021space}, which are computationally intensive and memory-demanding, necessitating large-scale, sophisticated GPUs for progress.
Interestingly, 2D image models on 2D super images created by rearranging video frames into multiple rows and columns~\cite{fan2022can} have proven to be a powerful machinery for video action recognition without being parameter heavy.
Naturally, recasting video action recognition as 2D image classification enables a plethora of highly efficient self-supervised learning approaches designed for images to be equally applicable for scalable and efficient self-supervised video representation learning.

In this work, we propose \underline{\textbf{Video}} \underline{\textbf{M}}asked \underline{\textbf{S}}iamese \underline{\textbf{N}}etwork (\ours) that masks super images created from unlabeled videos and leverages a masked Siamese network for invariant representation learning across masked and original views of the super images.
Given a super image created by rearranging a sequence of input video frames into multiple rows and columns, \ours~randomly masks patches from one view while leaving the other view unchanged.
To avoid temporal redundancy across frames, we make sure that if a region is masked in any frame, the same region is also masked across all the frames avoiding information leakage~\cite{tong2022videomae}.
Masking in images helps to get a strong encoder by breaking the spatial continuity of the images.
We conjecture that the complexity in videos due to the additional temporal dynamics demands breaking the temporal structure and forces the encoder to be invariant to this.
Thus we propose a more aggressive \textit{temporal masking} by dropping whole frames and the encoder learns to match the representations of an unmasked super image and super images with dropped frames.
Unlike reconstruction-based methods, \ours~eliminates the need for pixel-level reconstruction making the framework decoder-free.
An additional benefit of our decoder-free design is the ability to leverage off-the-shelf image encoders (\textit{e.g.}, ViT) pretrained solely on image data, thereby significantly reducing the amount of self-supervised pretraining with video data.
In contrast, encoder-decoder based MIM approaches typically require training from scratch due to the lack of image-pretrained decoders which results in longer pretraining schedules.

By leveraging a 2D image classification pipeline for video understanding and incorporating view-invariant masked image modeling, our approach is highly efficient across multiple dimensions: it is parameter-efficient, GPU memory friendly and label-efficient.
It reaches state-of-the-art performance with significantly fewer self-supervised pretraining epochs with video data.
Fig. \ref{fig:teaser} shows a comparison of the number of self-supervised pretraining epochs of different approaches along with video classification accuracy.
This shows that the compute-heavy pretraining with video data for \ours~can be upto $160\times$ fewer epochs compared to state-of-the-art approaches like VideoMAE~\cite{tong2022videomae}, MGM~\cite{fan2023motion} or MME~\cite{sun2023masked}, while maintaining the performance on standard video benchmarks.
We perform extensive experiments on four benchmark datasets and demonstrate the superiority of \ours~over contemporary self-supervised video action recognition approaches.
\newline
\\
\noindent Our key contributions are as follows:
\begin{itemize}
  \item To the best of our knowledge, \ours~is the first work that successfully extends the masked Siamese network to videos leveraging an image classification pipeline for self-supervised video representation learning.
  \item In addition to dropping spatial patches from frames, we show that the strategy of dropping or masking frames completely in the temporal dimension to create the masked view of the videos results in better representation learning.
  \item Our parameter-efficient encoder-only design enables self-supervised adaptation of image-pretrained vision transformers to videos, requiring up to $160\times$ fewer additional video pretraining epochs than recent generative masked video modeling approaches while remaining competitive or superior on benchmark datasets.
\end{itemize}
\section{Related Work}
\noindent\textbf{2D Action Recognition.} Since videos contain information along both spatial and temporal dimensions, spatio-temporal or 3D processing for recognizing actions has been the mainstay for a long time.
However, 2D image models have also been used for action recognition to enhance memory and compute efficiency.
Early approaches~\cite{zhao2017single,safaei2019still} used a single image from videos to recognize actions.
Later works create a representative image from the video by informative frame synthesis~\cite{qiu2021condensing}, adaptive spatio-temporal distillation~\cite{tavakolian2019awsd} and adversarial video distillation~\cite{tavakolian2019avd}.
However, using a single image to represent actions is limiting and hurts the performance.
Several later works~\cite{wang2016temporal,zhou2018temporal,lin2019tsm,Fan2019Tam} used different aggregation modules on 2D image backbones.
TSM~\cite{lin2019tsm} and its improvement TAM~\cite{Fan2019Tam} shift channels of 2D-CNNs along the temporal dimension.
Another set of approaches predicts actions by identifying key frames of an activity~\cite{wu2019adaframe, meng2020ar, sun2021dynamic}.
With the success of vision transformers, action recognition frameworks started exploring them~\cite{neimark2021video,fan2021multiscale,zhang2021vidtr}.
Recently, action recognition in videos is cast as an image classification problem~\cite{fan2022can, iqbal2024sitar} in which the frames are combined in a spatial grid to form a super image and classified using a Swin Transformer for images~\cite{liu2021swin}.

\noindent\textbf{Self-supervised Video Representation Learning.}
These approaches, being supervised, are critically dependent on large datasets requiring labels.
Self-supervised representation learning models address this by leveraging unlabeled data. These approaches can be categorized into three broad paradigms.
The first, transformation prediction, uses pretext tasks such as solving space-time puzzles~\cite{jing2018self,kim2019self}, predicting clip order~\cite{misra2016shuffle,luo2020video,fernando2017self} or estimating playback speed~\cite{benaim2020speednet, cho2021self}.
The second, contrastive learning, trains the model to align different augmentations of the same clip while separating others~\cite{pan2021videomoco, feichtenhofer2021large, ranasinghe2022self}.
The third, masked video modeling~\cite{fan2023motion, huang2023mgmae, sun2023masked}, adopts a mask-and-predict framework, where models like VideoMAE~\cite{tong2022videomae} mask a large portion of video tokens and reconstruct them using an encoder-decoder architecture.
However, such reconstruction-based methods try to capture unnecessary visual details and incur high computational cost.
While SMILE~\cite{thoker2025smile} attempts to improve semantic learning by predicting the CLIP features corresponding to synthetic motion, it still relies on an encoder-decoder architecture.
\ours, on the other hand, eliminates the decoder entirely.
Our approach replaces pixel reconstruction with feature alignment between masked and unmasked super images, using clustering losses with learnable prototypes and entropy maximization.
This decoder-free design retains high-level spatio-temporal semantics and substantially reduces pretraining epochs with videos.

\noindent\textbf{Masked Input Modeling for Vision Transformers.}
The idea of reconstructing masked inputs as a generative pretext task originated with denoising autoencoders~\cite{pathak2016context} and was later scaled to vision transformers through MAE~\cite{he2022masked}, which showed that masked reconstruction can produce highly transferable representations.
Building on this, various image-based methods~\cite{he2022masked, nguyen2023r, wang2023masked} have advanced masked image modeling, while approaches like JEPA~\cite{assran2023self} shifted the objective to predicting masked tokens in latent space, achieving strong performance.
In videos, VideoMAE~\cite{tong2022videomae} introduced masked autoencoding, prompting follow-up works to refine reconstruction targets~\cite{wang2023masked, wang2022bevt} and incorporate motion-aware masking strategies~\cite{sun2023masked}.
These masked autoencoders reconstruct original signals from corrupted inputs through an encoder-decoder setup.
Parallel to this, Siamese networks~\cite{10.5555/2987189.2987282} have enabled robust representation learning through contrastive objectives~\cite{he2020momentum, caron2021emerging} and have recently been explored in conjunction with masked modeling~\cite{zhou2021ibot, assran2022masked}. However, no prior work has explored an asymmetric, decoder-free masked Siamese design for videos.
In this paper, we propose \ours, the first adaptation of Masked Siamese Networks for videos.
We position \ours as a highly efficient integration of super image representations, image-pretrained Vision Transformers, and Masked Siamese Learning, rather than a fundamentally new architectural paradigm.
\section{Methodology}
\label{sec:method}

In this section, we briefly revisit the Masked Siamese Network~\cite{assran2022masked} (MSN) used in images.
Then, we describe \ours~and its components in detail.

\subsection{Preliminaries}
\noindent\textbf{MSN}~\cite{assran2022masked} is a self-supervised learning framework introduced for image representation learning using a discriminative mask-denoising process.
MSN combines masked image modeling with a Siamese learning objective.
The method constructs two augmented views of the same image: one is partially masked (the \emph{anchor view}), and the other is unmasked (the \emph{target view}).
Both views are passed through a shared ViT encoder.
The model is trained to align the representation of the masked view with that of the unmasked view using a soft-distribution over a set of prototypes for both the anchor and target views.
This encourages the network to produce semantically meaningful features from incomplete visual inputs.
MSN achieves strong performance by learning visual representations without labels.

\subsection{Masking Strategy in \ours}

\begin{figure}[t]
  \centering
  \includegraphics[width=\columnwidth]{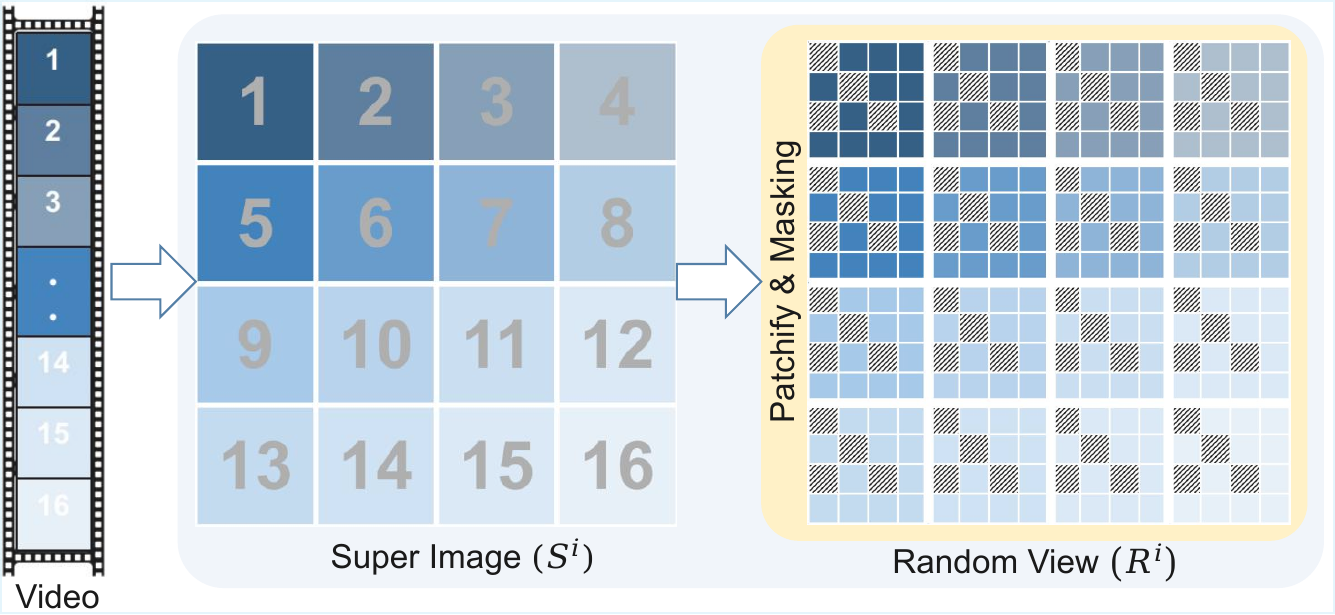}
  \vspace{-4pt}
  \caption{\textbf{Creation of super image and a random view.} Starting from an input video, a \textit{super image} \( S^i \) is formed by arranging \( M \) sampled frames into a 2D grid in a row-major format. Each frame is then divided into non-overlapping patches. A random binary mask is applied such that patches at the same spatial location across all frames are masked or retained together enforcing temporal consistency during masking. This results in the Random View ($R^i$). Masked patches are shown in gray for illustration. In practice, the token from that patch is dropped and not passed through the encoder.}
  \label{fig:random_view}
\end{figure}

Let $\mathcal{D}_u = \{ U^i \}_{i=1}^{N_u}$ be a collection of $N_u$ unlabeled videos and let $B$ be the number of such videos in each mini-batch during pretraining.
For each video $U^i$, we sample $M$ equidistant frames from non-overlapping temporal segments~\cite{wang2016temporal}.
Following SIFAR~\cite{fan2022can}, these frames are arranged in a fixed grid layout to form a 2D super image \( S^i\) (ref. Fig.~\ref{fig:random_view}).
Each super image ($S^i$) gives one target view ($T^i$) and a set of anchor views ($A^i$).
Note that the target view does not go through any masking; however, it is patchified into a set of non-overlapping patches and augmented.
Anchor views, on the other hand, go through masking (detailed below) after similar patchification and augmentation.

Following the masking strategy proposed in MSN~\cite{assran2022masked}, an anchor view, in our case, comprises of \textit{random views} and \textit{focal views} coming from the super image.
A random view $R^i$ of a super image $S^i$ is a result of masking spatial patches from the super image.
Specifically, each frame in $S^i$ is first divided into $N\times N$ non-overlapping patches, resulting in a set of $N^2$ tokens for each frame.
A random binary mask is applied over the patches such that all patches at the same spatial location across different frames
% in the super image
are either simultaneously masked or retained.
This design is consistent with the temporal tube masking formulation proposed in~\cite{tong2022videomae} and is shown to be effective in avoiding shortcuts in masked modeling for videos.
An illustration of the masking process is shown in Fig.~\ref{fig:random_view}.

\begin{figure}[tb]
  \centering
  \includegraphics[width=\columnwidth]{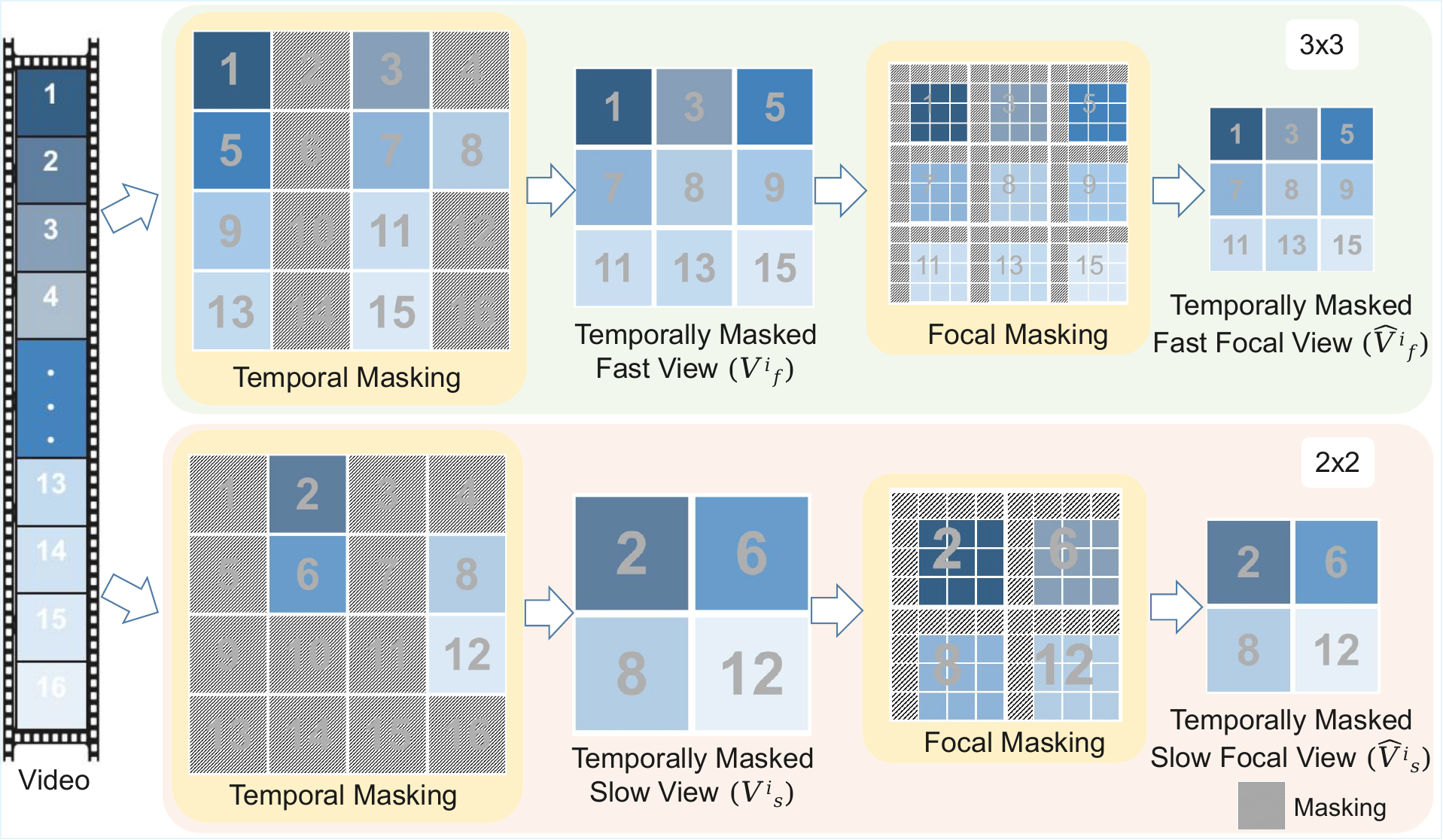}
  \vspace{-5pt}
  \caption{\textbf{Illustration of two-stage masking of focal views.} \textit{Temporal Masking} drops full frames to generate two different views: the \textit{Fast View} (\( V^i_f \)), retaining (approximately) half the frames of the target view of the super image and the \textit{Slow View} (\( V^i_s \)) containing approximately a quarter of the same. Subsequently, \textit{Focal Masking} is applied by selecting a spatially contiguous region and masking all surrounding patches, resulting in the \textit{Temporally Masked Fast Focal View} $\hat{V}^i_f$ and \textit{Temporally Masked Slow Focal View} $\hat{V}^i_s$. This hierarchical masking strategy helps the model attend to the informative regions in both space and time, enabling it to better capture motion patterns and spatial details across different action speeds.}
  \label{fig:anchor_view}
\end{figure}

To obtain a focal view ($F^i$) of a super image ($S^i$) we propose a two-stage masking mechanism.
The first stage is \emph{temporal masking} which is illustrated in Fig.~\ref{fig:anchor_view}.
It creates a temporally downsampled version of the super image by masking the whole of certain frames in the super image.
As a result, not all $M$ sampled frames are used in creating this view.
After this we create a \textit{fast view} and a \textit{slow view} from the frames that are retained.
\begin{itemize}
    \item \textbf{Fast View} $V^i_f$ is obtained by dropping approximately half the number of frames present in $S^i$, such that the number of frames forming $V^i_f$ is approximately $\frac{M}{2}$.
    \item \textbf{Slow View} $V^i_s$ is obtained by dropping an additional half the number of frames from $V^i_f$, making the number of frames in $V^i_s$ approximately $\frac{M}{4}$.\footnote{Slow/Fast nomenclature is motivated by SlowFast Networks~\cite{feichtenhofer2019slowfast}.}
\end{itemize}
The reason, the number of frames in the super image after temporal masking are kept \textit{approximately} $\frac{M}{2}$ and $\frac{M}{4}$, is that a super image works best when the layout is square \textit{i.e.}, equal number of rows and columns form the super image~\cite{fan2022can}.
Different from focal views in MSN applied to images, temporal masking in \ours~helps exploit different temporal sparsity of the same action, offering varied motion dynamics for the model to learn from.

The second stage involves \textit{focal masking} on each of these temporally masked views.
Specifically, a contiguous region in each frame
% of the super image
is randomly selected within each
% of the
temporally masked view and retained, while the surrounding patches are masked out.
Similar to the random views, we propose to mask out the same spatial area in each frame of the super image.
Following such a strategy, we get multiple focal views via multiple random selections of contiguous regions from each of the temporally masked views.
Mathematically, the fast view $V^i_f$ gives rise to $n$ temporally masked fast focal views $\{\hat{V}^i_{f,j}\}_{j=1}^n$.
Similarly, $n$ temporally masked slow focal views $\{\hat{V}^i_{s,j}\}_{j=1}^n$ are obtained from the slow view $V^i_s$.
The set of anchor views ($A^i$) contains random views, temporally masked fast focal views and temporally masked slow focal views.
This progressive masking strategy helps learn from both global context and localized high-resolution cues, improving spatio-temporal feature alignment across views of different temporal granularity.

\begin{figure}[tb]
  \centering
  \includegraphics[width=\columnwidth]{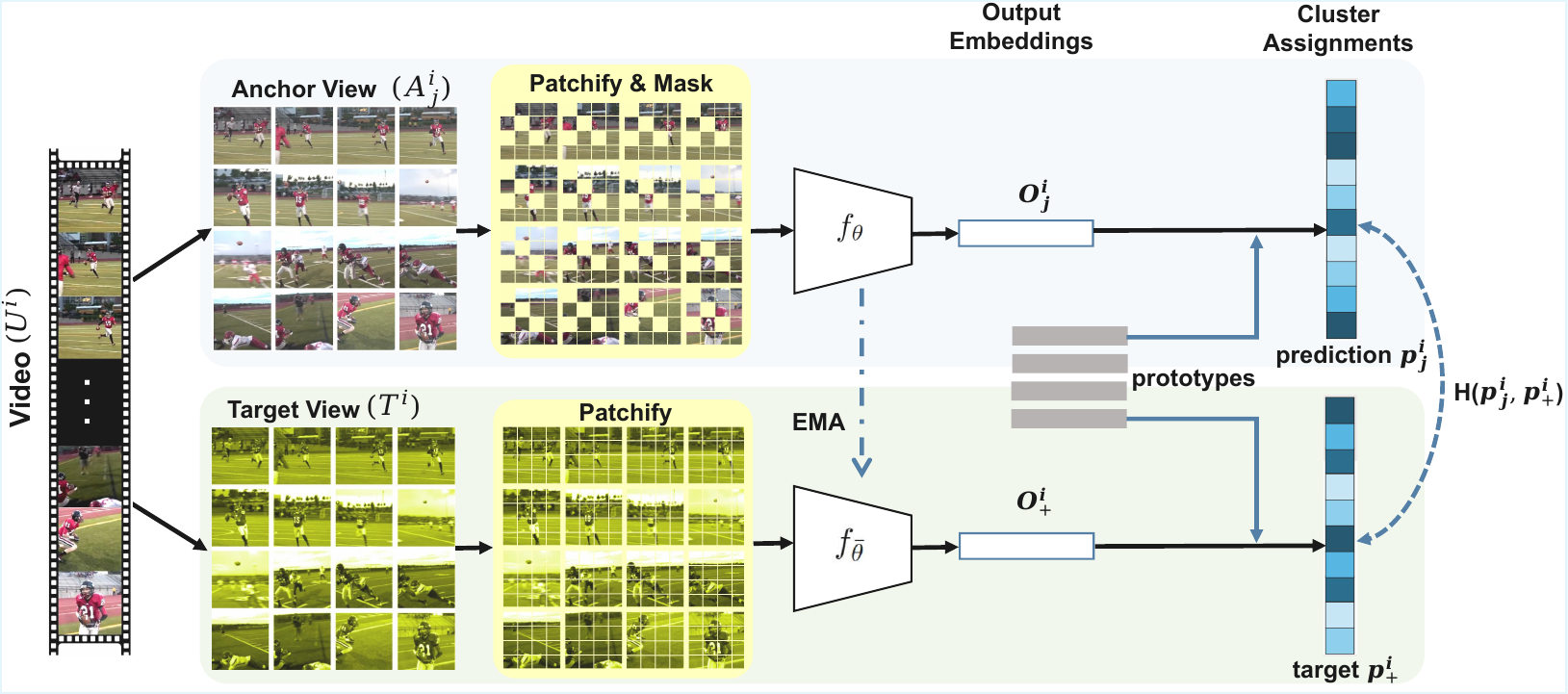}
  \vspace{-2mm}
    \caption{\textbf{Overview of \ours.} Given an unlabeled video clip $U^i$, equidistant frames are sampled and arranged into a 2D super image. Two different views are generated: the $j^{th}$ \textit{anchor view} $A^i_j$, which is randomly masked after patchification and the \textit{target view} $T^i$ which remains unmasked. Both views are processed using a Siamese architecture. $A^i_j$ is passed through the encoder $f_\theta(\cdot)$ to obtain embeddings $\mathbf{O}^i_j$, while $T^i$ is passed through the EMA updated target encoder $f_{\bar{\theta}}(\cdot)$ to produce $\mathbf{O}^i_+$. These representations are then assigned to cluster prototypes, generating a predicted distribution $\mathbf{p}^i_j$ for the anchor and a target distribution $\mathbf{p}^i_+$ for the target. The training objective is to align $\mathbf{p}^i_j$ with $\mathbf{p}^i_+$ using cross-entropy loss \(H(\mathbf{p}^i_j, \mathbf{p}^i_+)\), encouraging the masked anchor representation to match that of the unmasked target.
    }
  \label{fig:Main_model}
\end{figure}

\subsection{Learning in \ours}
Fig.~\ref{fig:Main_model} provides an overview of our \ours~approach.
Following clustering-based self-supervised learning frameworks~\cite{caron2020unsupervised, caron2021emerging, assran2021semi}, the super images corresponding to the target and the anchor views are passed through an encoder to produce feature representations, which are then projected onto a set of learnable prototypes.
The projections are converted into distributions and the encoder is encouraged to produce similar distributions coming from the masked anchor views and the unmasked target views using cross-entropy loss.

Let $f_\theta(\cdot)$ denote the parameterized anchor encoder and let $\mathbf{O}^i_j = f_\theta(A^i_j) \in \mathbb{R}^d$ represent the output embedding obtained from the $j^{th}$ anchor view $A^i_j$.
Note that the set of anchor views contains both random and temporally masked focal views.
Similarly, let $f_{\bar{\theta}}(\cdot)$ be the target encoder, parameterized by $\bar{\theta}$, and let $\mathbf{O}^i_+ = f_{\bar{\theta}}(T^i) \in \mathbb{R}^d$
denote the embedding computed from the target view $T^i$.
Following MSN~\cite{assran2022masked}, the target encoder weights $\bar{\theta}$ are updated using an exponential moving average (EMA) of the anchor encoder parameters $\theta$~\cite{grill2020bootstrap}.
Both encoders share the same ViT architecture~\cite{dosovitskiy2021an} and the final representation is taken from the [CLS] token at the output of the transformer.

\noindent\textbf{Prototype-driven Learning.} We adopt a prototype-based training objective to learn video representations without labels inspired by~\cite{assran2022masked}.
Specifically, we maintain a set of $P$ learnable prototypes each with dimension $d$.
The collection of prototype vectors are denoted as $\mathbf{q} \in \mathbb{R}^{P \times d}$.
We compute anchor and target predictions by measuring cosine similarity between the encoder outputs and the prototypes.
For the $j^{th}$ anchor representation $\mathbf{O}^i_j$, the prediction is computed as:
\vspace{-0.2cm}
\begin{equation}
\label{eq:prediction_pi}
\mathbf{p}^i_j = \mathrm{softmax}\left( \frac{\mathbf{q}\cdot \mathbf{O}^i_j}{\tau} \right),
\vspace{-0.2cm}
\end{equation}
with a temperature parameter \( \tau \in (0, 1) \), while the target prediction \( \mathbf{p}^i_+ \) is similarly obtained using a sharper temperature parameter \( \tau^+ < \tau \):
\vspace{-0.2cm}
\begin{equation}
\label{eq:prediction_p+}
\mathbf{p}^i_+ = \mathrm{softmax}\left( \frac{\mathbf{q}\cdot \mathbf{O}^i_+}{\tau^+} \right)\:.
\vspace{-0.2cm}
\end{equation}

The training objective consists of two components. First, we minimize the cross-entropy loss \(H \), between anchor prediction $\mathbf{p}^i_{j}$ and target prediction $\mathbf{p}^i_+$:
\vspace{-0.2cm}
\begin{equation}
\label{eq:cross_ent}
\mathcal{L}_{\text{align}} = \frac{1}{KB} \sum_{i=1}^{B} \sum_{j=1}^{K} H(\mathbf{p}^i_{j}, \mathbf{p}^i_+),
\vspace{-0.2cm}
\end{equation}
where $K$ denotes the total number of anchor views including random and all focal views.
We include a Mean Entropy Maximization (ME-MAX) regularizer~\cite{assran2021semi, joulin2012convex} to encourage uniform prototype usage.
We first, compute the average prediction $\bar{\mathbf{p}}$ across all the anchor views in the batch as,

\vspace{-0.2cm}
\begin{equation}
% \label{eq:me_max}
 \bar{\mathbf{p}} := \frac{1}{KB} \sum_{i=1}^{B} \sum_{j=1}^{K} \mathbf{p}^i_{j},
 \vspace{-0.2cm}
\end{equation}

\noindent and subsequently add the ME-MAX regularizer making the overall objective,
\vspace{-0.2cm}
\begin{equation}
% \label{eq:me_max}
\mathcal{L}_{\text{total}} = \mathcal{L}_{\text{align}} - \lambda H(\bar{\mathbf{p}}),
\vspace{-0.2cm}
\end{equation}
where $\lambda$ is the weight of the regularizer.
This joint objective has been shown to prevent representation collapse while promoting discriminative embeddings~\cite{assran2021semi, joulin2012convex}.
\section{Experiments}
\label{sec:exp}

In this section, we present comparative evaluation of our \ours~framework.
We also perform comprehensive ablation studies to verify the effectiveness of different components and hyperparameter sweep to choose crucial hyperparameters.

\noindent \textbf{Datasets and Backbones.}
We evaluate \ours~on four widely used datasets: Kinetics-400~\cite{kay2017kinetics}, UCF101~\cite{soomro2012ucf101}, HMDB51~\cite{hmdb} and Something-Something V2 (SSV2)~\cite{goyal2017something}.
As the backbone to our encoder, we use the small and base versions of Vision Transformer~\cite{dosovitskiy2021an} (denoted as ViT-S and ViT-B respectively).
We used the pretrained DeiT-v3 and DINO-v3-distilled checkpoints made available by the authors of~\cite{touvron2022deit} and~\cite{simeoni2025dinov3} respectively to initialize our encoder.
These two variations are denoted as \ours-DeiT and \ours-DINO respectively.

\noindent \textbf{Implementation Details.}
Our pre-training pipeline follows~\cite{fan2022can}.
We adopt uniform frame sampling and apply standard multi-scale jittering followed by Gaussian blur.
The frames are randomly cropped to a resolution of $224\times 224$ before forming the super images.
Consistent with~\cite{assran2022masked}, we set the number of prototypes $P$ to $1024$ with dimension $d = 256$.
The hyperparameters $\tau, \tau^+$ and $\lambda$ are set to $0.1, 0.025$ and $5.0$ respectively.
The number of focal views is set to $6$ unless otherwise specified.
Half of the focal views are fast and the rest are slow focal views.
We use a masking ratio of $0.7$ \textit{i.e.}, $70\%$ of tokens are dropped while creating the anchor views, except for SSV2 where the value is taken to be $0.5$.
Drop Path and weight decay values are kept both $0.01$.
All experiments were conducted on a server with 4 NVIDIA H100 GPUs.
All \ours-DeiT/\ours-DINO models undergo the proposed self-supervised pretraining for $50/10$ epochs, including a $5/1$ epoch linear warm-up.
We used AdamW optimizer~\cite{loshchilov2017decoupled} and follow a cosine learning rate.

For full fine-tuning, we used Mixup~\cite{zhang2018mixup} and CutMix~\cite{9008296} with mixing coefficients of $0.8$ and $1.0$, respectively and apply label smoothing with coefficient $0.1$.
All models are fine-tuned for 30 epochs unless otherwise mentioned.
We used AdamW optimizer with a weight decay of $0.01$ and follow a cosine learning rate scheduler for all datasets.
For evaluation, we used $5~clips \times 3~crops$ for Kinetics-400 and UCF101, $10~clips \times 3~crops$ for HMDB51 and $2~clips \times 3~crops$ setup for SSV2.
Performances are shown in terms of Top-1 accuracy averaged across 2 random seeds, unless otherwise mentioned.

\noindent \textbf{Choosing hyperparameters.}
We identified some crucial hyperparameters by conducting a sweep on a representative subset of the Kinetics-400 training set, created by randomly sub-sampling $25\%$ of the data per class.
The configuration that maximized performance on the full validation set was subsequently adopted for all experimental benchmarks.
More details of this analysis are provided in the Appendix.

\noindent \textbf{Comparison.}
We compare against state-of-the-art MAE based approaches like VideoMAE \cite{tong2022videomae} and its architectural variants as well as very recent approaches like SIGMA-DINO~\cite{salehi2024sigma}, SMILE~\cite{thoker2025smile} \textit{etc}.
While VideoMAEv2 \cite{wang2023videomae} introduces additional improvements, it relies on significantly larger backbones and heavy distillation from large teacher models, requiring computational resources beyond our scope and thus we do not directly compare with this in our setting.
Therefore, our comparisons focus on methods operating under efficient, distillation-free pretraining settings.
SMILE~\cite{thoker2025smile} relies on additional synthetic motion signals beyond unlabeled videos.
Importantly, since our formulation leverages super images rather than video data, we retain the standard 2D patch embedding without any 3D inflation.
This results in a reduction of parameters, as shown in Table~\ref{tab:k400}.
\begin{table}[t!]
\small
\centering

\resizebox{\textwidth}{!}{
\begin{tabular}{l|c|c|r|r|c}

\multirow{2}{*}{\textbf{Method}} & \multirow{2}{*}{\textbf{Backbone}} & \multirow{2}{*}{\textbf{Decoder}} & \multirow{2}{*}{\textbf{Epochs}} & \textbf{Params} & \multirow{2}{*}{\textbf{Top-1 ($\uparrow$)}}\\
&&&&\textbf{(M $\downarrow$)} & \\ \hline

VideoMAE~\cite{tong2022videomae} \small{(\texttt{NeurIPS'22})} & ViT-S & \checkmark & 800 & 22 & 79.0 \\
SIGMA-DINO~\cite{salehi2024sigma} \small{(\texttt{ECCV'24})} & ViT-S & \checkmark & 800 & 22 & 79.4\\

SMILE (motion)~\cite{thoker2025smile} \small{(\texttt{CVPR'25})} & ViT-S & \checkmark & 800 & 22 & 79.5\\ % 

\cellcolor{cyan!8}\ours-DeiT (Ours) & \cellcolor{cyan!8}ViT-S & \cellcolor{cyan!8}\xmark & \cellcolor{cyan!8}50 & \cellcolor{cyan!8}21.5 & \cellcolor{cyan!8}80.0 \\ 
\cellcolor{cyan!8}\ours-DINO (Ours) & \cellcolor{cyan!8}ViT-S & \cellcolor{cyan!8}\xmark & \cellcolor{cyan!8}10 & \cellcolor{cyan!8}21 & \cellcolor{cyan!8}\textbf{80.8} \\ 
\shline

SVT~\cite{ranasinghe2022self} \small{(\texttt{CVPR'22})} & ViT-B & \xmark & 20 & 121 & 78.1 \\
MGM~\cite{fan2023motion} \small{(\texttt{ICCV'23})} & ViT-B & \checkmark & 800 & 87 & 80.8 \\
CMAE-V~\cite{lu2023cmae} \small{(\texttt{ArXiv'23})} & ViT-B & \checkmark & 800 & 87 & 80.2 \\
MGMAE~\cite{huang2023mgmae} \small{(\texttt{CVPR'23})} & ViT-B & \checkmark & 800 & 87 & 81.2 \\
OmniMAE~\cite{girdhar2023omnimae} \small{(\texttt{CVPR'23})} & ViT-B & \checkmark & 800 & 87 & 80.8 \\
ViC-MAE~\cite{hernandez2024vic} \small{(\texttt{ECCV'24})} & ViT-B & \checkmark & 800 & 87 & 80.8\\
ST-MAE~\cite{feichtenhofer2022masked} \small{(\texttt{NeurIPS'22})} & ViT-B & \checkmark & 800 & 87 & 81.3\\
VideoMAE~\cite{tong2022videomae} \small{(\texttt{NeurIPS'22})} & ViT-B & \checkmark & 800 & 87 & 80.0 \\
SIGMA-DINO~\cite{salehi2024sigma} \small{(\texttt{ECCV'24})} & ViT-B & \checkmark & 800 & 87 & 81.6 \\
VideoMAE~\cite{tong2022videomae} \small{(\texttt{NeurIPS'22})} & ViT-B & \checkmark & 1600 & 87 & 81.5 \\ % \cdashlinelr{1-6}
MGM~\cite{fan2023motion} \small{(\texttt{ICCV'23})} & ViT-B & \checkmark & 1600 & 87 & 81.7 \\ 
MME~\cite{sun2023masked} \small{(\texttt{CVPR'23})} & ViT-B & \checkmark & 1600 & 87 & 81.8 \\ % \cdashlinelr{1-6}
SMILE (motion)~\cite{thoker2025smile} \small{(\texttt{CVPR'25})} & ViT-B & \checkmark & 600 & 87 & 83.1 \\ % \cdashlinelr{1-6}

\cellcolor{cyan!8}\ours-DeiT (Ours) & \cellcolor{cyan!8}ViT-B & \cellcolor{cyan!8}\xmark & \cellcolor{cyan!8}50 & \cellcolor{cyan!8}86.5 & \cellcolor{cyan!8}82.0\\
\cellcolor{cyan!8}\ours-DINO (Ours) & \cellcolor{cyan!8}ViT-B & \cellcolor{cyan!8}\xmark & \cellcolor{cyan!8}10 & \cellcolor{cyan!8}86 & \cellcolor{cyan!8}{\textbf{83.3}}\\  \shline
\end{tabular}}

\vspace{3mm} 
\caption{\textbf{Comparison on Kinetics-400.} Our \ours is initialized from pretrained image ViT weights followed by pretraining and fine-tuning on Kinetics-400. We achieve superior top-1 accuracy in both \ours-DeiT and \ours-DINO after pretraining only for 50 and 10 epochs respectively. \textbf{Note:} The parameter counts for the competing methods are reported from the respective papers and they correspond to the learnable parameters in the encoder. However, most of these methods employ an additional decoder during pretraining, increasing actual training parameters. In contrast, our approach is decoder-free.}
\label{tab:k400}
\normalsize
\end{table}

\subsection{Experimental Results and Analysis}

\noindent\textbf{Results on Kinetics-400.} We conduct self-supervised pretraining on Kinetics-400 for only 50 and 10 epochs in case of \ours-DeiT and \ours-DINO respectively.
Remarkably, \ours-DeiT (ViT-S) surpasses the second best model by $0.5\%$ while \ours-DINO surpasses it by $1.3\%$.
For ViT-B architecture, our best model \ours-DINO improves the state-of-the-art by $0.2\%$.
It is worth noting that the second best approach SMILE requires 600 epochs of pretraining (a $60\times$ increase) yet our performance remains superior.
For some of the MAE based approaches \textit{e.g.}, VideoMAE, MGM and MME, the increase in pretraining epochs is $160\times$.
We attribute this substantial efficiency gain to the MSN architecture of \ours, which, unlike encoder-decoder frameworks, eliminates the need for training uninitialized decoder weights from scratch and relies less on low-level detail learning for reconstruction.
To compare, we initialize the ViT-B encoder of VideoMAE with ImageNet-21K pretrained weights like ours and run pretraining for $50$ epochs followed by $30$ epochs of finetuning on Kinetics-400.
The model, not surprisingly, yields only 57\% top-1 accuracy.
One possible reason is that such a limited amount of video pretraining may be insufficient for effectively training the randomly initialized decoder.
This observation suggests that our decoder-free design is better suited to efficiently adapt strong image-pretrained encoders to video representation learning under short video training schedules.
The tendency of MAEs to emphasize on unnecessary low-level details may necessitate longer training.

\begin{table}[tb]

\resizebox{\textwidth}{!}{
\begin{tabular}{lccccc}
Dataset & Backbone  &  MoCo v3  & VideoMAE & \cellcolor{cyan!8} \ours-DeiT & \cellcolor{cyan!8} \ours-DINO \\
\shline 
UCF101 & ViT-B & 81.7 & 91.3 \macc{3200 ep} & \cellcolor{cyan!8} 92.0 \pacc{50 ep} & \cellcolor{cyan!8} \textbf{95.4} \pacc{10 ep} \\

HMDB51 & ViT-B & 39.2 & 62.6 \macc{4800 ep} & \cellcolor{cyan!8} 62.6 \pacc{50 ep} & \cellcolor{cyan!8} \textbf{70.4} \pacc{10 ep} \\ 

SSV2 & ViT-S & - & \textbf{66.8} \macc{2400 ep} & \cellcolor{cyan!8} {64.8} \pacc{50 ep} & \cellcolor{cyan!8} 65.8 \pacc{10 ep} \\ 

SSV2 & ViT-B & 54.2 & \textbf{70.8} \macc{2400 ep} & \cellcolor{cyan!8} {69.0} \pacc{50 ep} & \cellcolor{cyan!8} 69.4 \pacc{10 ep} \\ 
\shline
\end{tabular}}
\vspace{3mm}
\caption{Comparisons with the results of previous \textbf{self-supervised pre-training methods} on UCF101, HMDB51, and SSV2, using 16-frame inputs and ViT-S/ViT-B backbones. All methods utilize unlabeled training data for pre-training and consider the labels only for fine-tuning. \ours~ delivers significantly improved performance achieving state-of-the-art results on UCF101 and HMDB51. While slightly trailing on SSV2, our approach offers immense efficiency, very less pretraining epochs across all the datasets compared to VideoMAE. The values reported in this table use a single fixed seed due to computational constraints.}
\label{tab:ssv2}
\end{table}

\vspace{2mm}
\noindent\textbf{Results on UCF101 and HMDB51.} As shown in Table~\ref{tab:ssv2}, our proposed \ours~consistently achieves superior top-1 recognition accuracy compared to other self-supervised frameworks such as MoCo v3~\cite{9711302} and VideoMAE on small-scale datasets like UCF101 and HMDB51.
Remarkably, while prior methods rely on prohibitively long pretraining schedules (3200 and 4800 epochs for UCF101 and HMDB51, respectively), \ours-DINO attains higher performance with only 10 epochs of pretraining, yielding a $320\times$ reduction in training epochs on UCF101 and $480\times$ on HMDB51.
This result underscores the architectural efficiency and inductive strength of our framework without requiring extensive compute.
Such training efficiency is particularly advantageous in regimes with limited labeled data or constrained computational budgets.

\vspace{2mm}
\noindent\textbf{Results on SSV2.} As shown in Table~\ref{tab:ssv2}, \ours-DINO lags behind VideoMAE by a small margin (1.0\% for ViT-S and 1.4\% for ViT-B).
We contend that this is a direct and well-justified trade-off for our framework's superior computational efficiency.
This performance is achieved using $240\times$ fewer pre-training epochs for \ours-DINO.
Achieving competitive results on a difficult, motion-centric benchmark like SSV2 with a fraction of the computational budget highlights the superior scalability and practical utility of our approach.

\subsection{Low-shot classification}
In this section, we show the generalizability of the proposed self-supervised video representation learning approach.
Following established protocols~\cite{salehi2024sigma, thoker2025smile}, we chose the ViT-B variant of \ours-DeiT and \ours-DINO pretrained on $16$ frame input from Kinetics-400 and perform the \textbf{Low-shot classification} experiment. Here, we evaluate the learned representation on action recognition with few training samples per-category.
We follow the setup in~\cite{thoker2025smile} and finetune with a total of 1000 training examples randomly sampled from UCF101.
Table~\ref{tab:severe} shows the results, in which our method substantially outperforms the rest, demonstrating its strong few-shot action recognition capability in videos.

\begin{table}[t]
\centering
\setlength{\tabcolsep}{4pt}

\begin{tabular}{lcccccc}
\toprule
Dataset 
& VideoMAE 
& MME 
& SIGMA 
& SMILE
& \cellcolor{cyan!8}  \ours 
& \cellcolor{cyan!8}  \ours\\

& \small{\texttt{NeurIPS'22}}
& \small{\texttt{CVPR'23}}
& \small{\texttt{ECCV'24}}
& \small{\texttt{CVPR'25}}
& \cellcolor{cyan!8}  \small{DeiT}
& \cellcolor{cyan!8}  \small{DINO}\\

\midrule

UCF101 
& 74.6 
& 79.2 
& 84.1 
& 86.4 
& \cellcolor{cyan!8}  88.0 
& \cellcolor{cyan!8}  \textbf{89.0} \\

\bottomrule
\end{tabular}
\vspace{4mm}
\caption{Comparison of low-shot action recognition on UCF101 using only $1000$ training videos, 16-frame inputs and ViT-B backbones for fine-tuning. Despite using substantially fewer video pretraining epochs, both \ours-DeiT and \ours-DINO outperform prior approaches, with \ours-DINO achieving the best top-1 accuracy of $89.0\%$, improving over SMILE by $2.6\%$. The results highlight the strong transferability and label efficiency of the representations learned by \ours{}.}
\label{tab:severe}

\end{table}

\subsection{Additional Experiments and Ablation Studies}
In this subsection, we first provide additional analysis to better understand the contribution of the proposed \ours pretraining on top of strong image-pretrained initialization. 
We then present comprehensive ablation studies to validate the effectiveness of the different design choices in \ours. 
Unless otherwise specified, all experiments are conducted using \ours-DeiT with a ViT-S backbone on Kinetics-400 with 16-frame input. 
The pretrained models are fine-tuned for 30 epochs and evaluated using a consistent inference protocol of $5$ clips $\times$ $3$ crops.

\begin{table*}[t]
\centering
\begin{tabular}{lcccc}
Method & DeiT-S & DeiT-B & DINO-S & DINO-B \\
\midrule
w/o VideoMSN & 77.3 & 78.2 & 80.7 & 83.1 \\
\cellcolor{cyan!8} w/ VideoMSN &
\cellcolor{cyan!8}\textbf{80.0} &
\cellcolor{cyan!8}\textbf{82.0} &
\cellcolor{cyan!8}\textbf{80.8} &
\cellcolor{cyan!8}\textbf{83.3} \\
\bottomrule
\end{tabular}
\vspace{4mm}
\caption{\textbf{Effect of \ours pretraining on Kinetics-400.} All models are initialized from the same image-pretrained DeiT-v3 or DINO-v3 checkpoints. ``w/o \ours'' denotes direct fine-tuning of the image-pretrained backbone for action recognition, whereas ``w/ \ours'' first performs the proposed lightweight self-supervised video pretraining stage before fine-tuning. \ours consistently improves performance across all backbones, with particularly large gains for DeiT-S and DeiT-B (+2.7\% and +3.8\% top-1 accuracy, respectively), while also providing modest improvements for the already strong DINO-S and DINO-B initializations.}
\label{tab:additional}
\end{table*}
\paragraph{Effect of \ours{} Pretraining over Image-pretrained Initialization.}
We analyze the impact of the proposed \ours{} pretraining by starting from strong image-pretrained DeiT-v3 and DINO-v3 checkpoints and evaluating two training settings on Kinetics-400.
In the first setting (\textit{w/o} \ours), the image-pretrained backbone is directly fine-tuned for action recognition without any self-supervised video adaptation.
In the second setting (\textit{w/} \ours), we first perform the proposed \ours{} self-supervised video pretraining stage before fine-tuning on Kinetics-400.
As shown in Table~\ref{tab:additional}, \ours{} consistently improves performance across all image-pretrained initializations.
The gains are particularly pronounced for DeiT-S and DeiT-B, yielding improvements of $+2.7\%$ and $+3.8\%$ top-1 accuracy, respectively.
Even for the stronger DINO-v3 initialization, \ours{} provides additional improvements, demonstrating that lightweight self-supervised video adaptation can effectively enhance image-pretrained representations for spatio-temporal video understanding and help in the case where adaptation in a short time is necessary.

\begin{table*}[htb]
\centering
\begin{tabular}{lccc}
\toprule
Method & Epochs ($\downarrow$) & Wall-clock Time ($\downarrow$) & Total FLOPs ($\downarrow$) \\
\midrule
VideoMAE & 1600 & 266.7 h & 40.89 E \\
\cellcolor{cyan!8} \ours-DINO (ours) & \cellcolor{cyan!8} \textbf{10 } & \cellcolor{cyan!8} \textbf{21.7 h} & \cellcolor{cyan!8} \textbf{4.58 E} \\
\midrule
Reduction & \textbf{160$\times$} & \textbf{12.3$\times$} & \textbf{8.9$\times$} \\
\bottomrule
\end{tabular}
\vspace{3mm}
\caption{Total pretraining epochs, wall-clock time, and FLOPs for VideoMAE and \ours-DINO on Kinetics-400, showing the substantially lower overall training cost of \ours.}
\label{tab:flop_time}
\end{table*}

\noindent\textbf{Training Time Compute.} To better quantify efficiency of \ours, we report the total video pretraining time and total FLOPs in Table~\ref{tab:flop_time} using the same hardware (4$\!\times\!$H100 GPUs) for VideoMAE and \ours.
The wall-clock time is measured as \# epochs (ep) $\!\times\!$ time/epoch (t/ep) giving 266.7 hours for VideoMAE (1600 ep$\!\times\!$10 min/ep) and 21.7 hours for \ours-DINO (10 ep$\!\times\!$130 min/ep); a reduction in video pretraining time by 12.3$\!\times\!$.
Total FLOPs is given by GFLOP/sample $\times$ sample/ep $\times$ ep.
VideoMAE requires $(106.5 \!\times\! 240\mathrm{K} \!\times\! 1600)\!\approx\!40.89$ EFLOPs, whereas \ours-DINO requires $(1911.9 \!\times\! 240\mathrm{K} \!\times\! 10)\!\approx\!4.58$ EFLOPs implying a reduction of $\!\sim\! 8.9\!\times\!$ in total FLOPs.
Although \ours~has higher FLOPs per epoch,
the substantially shorter video pretraining schedule lowers the total computation.

\begin{table*}[t]
\centering
\captionsetup[table]{labelformat=empty}
\begin{minipage}[t]{0.45\linewidth}
\centering
\begin{tabular}{lx{24}}
Approach & Top-1\\
\shline
w/o Temp Aug & 78.6 \\
w/ Temp Aug  & \cellcolor{cyan!8} 80.0 \\
\end{tabular}
\vspace{2mm}
\captionof{table}{\textbf{(a) Temporal Augmentations:} We evaluate the impact of temporal masking on representation learning. Incorporating temporal augmentations by using 3×3 and 2×2 focal view improves top-1 accuracy by 1.4 points over the
baseline without augmentation.}
\addtocounter{table}{-1}
\sublabel{tab:temporal_aug}{a}
\end{minipage}
\hfill
\begin{minipage}[t]{0.45\linewidth}
\centering
\begin{tabular}{lx{24}}
Approach & Top-1\\
\shline
w/o Sinkhorn & \cellcolor{cyan!8} 80.0 \\
w/ Sinkhorn  & 79.1 \\
\end{tabular}
\vspace{2mm}
\captionof{table}{\textbf{(b) Sinkhorn Normalization:} We analyze the impact of Sinkhorn normalization and observe that excluding it leads to a 0.9\% gain in top-1 accuracy, indicating that the ME-MAX regularizer alone is sufficient to preserve representation diversity in our model.}
\addtocounter{table}{-1}
\sublabel{tab:sinkhorn}{b}
\end{minipage}
\medskip

\vspace{0.1mm}
\begin{minipage}[t]{0.45\linewidth}
\centering
\begin{tabular}{x{80}x{30}}
Anchor Views & Top-1 \\
\shline
Only Random View & 78.5 \\
Only Focal View  & 77.1 \\
Both Views       & \cellcolor{cyan!8}{80.0} \\
\end{tabular}
\vspace{2mm}
\captionof{table}{\textbf{(c) Anchor View:} `Both Views' (Random and Focal) perform better, achieving the optimal 80.0\% accuracy. `Both Views' outperforms using `Only Random View' (78.5\%) or `Only Focal View' (77.1\%) in isolation, validating our multi-view masking design.}
\addtocounter{table}{-1}
\sublabel{tab:effect_anchor_view}{c}
\end{minipage}
\hfill
\begin{minipage}[t]{0.45\linewidth}
\centering
\begin{tabular}{lx{30}}
Focal Views & Top-1 \\
\shline
Only Fast Views & 78.6 \\
Only Slow Views & 77.5 \\
Both Views      & \cellcolor{cyan!8}{80.0} \\
\end{tabular}
\vspace{2mm}
\captionof{table}{\textbf{(d) Focal Views:} Use of both fast and slow views performs the best (80.0\%), which significantly outperforms only fast (78.6\%) or only slow views (77.5\%). It demonstrates the importance of training the model on varied temporal
sampling rates.}
\addtocounter{table}{-1}
\sublabel{tab:ablation_focal_view}{d}
\end{minipage}
\captionsetup[table]{labelformat=default}
\vspace{2mm}
\caption{\textbf{Ablation studies on Kinetics-400 with a 16-frame ViT-S backbone.} All models are pre-trained for 50 epochs and fully fine-tuned for evaluation. We adopt a consistent inference protocol of 5 clips × 3 crops. The default configuration (highlighted) achieves optimal performance across key components: (a) Temporal augmentations, (b) Sinkhorn normalization, (c) Anchor view, and (d) Focal views. The values reported in this table use a single fixed seed due to computational constraints.}
\label{tab:ablations}
\end{table*}

\noindent\textbf{Effect of Temporal Augmentations.} We examine how adding temporal augmentations affects the performance of our \ours-DeiT model.
Table~\ref{tab:temporal_aug} shows that incorporating temporal masking boosts accuracy by 1.4 percentage points compared to the variant that omits this augmentation.
As depicted in Fig.~\ref{fig:anchor_view}, temporal masking employs a hierarchical scheme to create six small focal views.
Starting with 16 uniformly sampled frames, we drop 7 frames to build $3\times3$ super images which act as the fast views.
We create 3 such fast views.
Similarly, we drop additional 5 frames (\textit{i.e.}, 12 frames are dropped altogether) to build $2\times2$ super images which act as the slow views.
We created 3 different views for this case also.
The outcome of this augmentation appears in the second row of Table~\ref{tab:temporal_aug}.
For the baseline without temporal augmentation, all 16 frames are retained for each of the six focal views.
All such focal views, in this baseline, are arranged as a $4\times 4$ super image. This configuration leads to a $1.4$ percentage point decrease in performance, as listed in the first row of Table~\ref{tab:temporal_aug}.
We attribute the improvement to the ability of the temporal augmentation to encourage the model to learn motion-aware features. 
Since $4\times 4$ focal views may dilute motion cues and action semantics by overly compressing frames, using smaller grids like $3\times 3$ and $2\times 2$ helps preserve and highlight the actions of the video.

\noindent\textbf{Effect of Sinkhorn Normalization.} For \ours, we follow the default setting of Masked Siamese Network and set the ME-MAX regularization weight $\lambda$ to $5.0$.
However, in this experiment, we explore the interaction between this regularization strength and Sinkhorn normalization.
To investigate this, we study the impact of Sinkhorn normalization during pretraining. Table~\ref{tab:sinkhorn} shows that omitting Sinkhorn while keeping $\lambda$ to $5.0$ produces the best performance. This result indicates that Sinkhorn normalization may interfere with optimal feature learning when used alongside strong regularization.

\noindent\textbf{Ablation on Anchor Views.} Table~\ref{tab:effect_anchor_view} compares the efficacy of our masking strategies using our \ours-DeiT (ViT-S). The results clearly show a synergistic effect: using only random views yields 78.5\% accuracy, and using only focal views results in 77.1\%. The combination of both random and focal views, as used in our full model, produces the highest Top-1 accuracy of 80.0\%.

\noindent\textbf{Ablation on Focal Views.} Table~\ref{tab:ablation_focal_view} evaluates the impact of our temporal masking strategy.
The results demonstrate that using both Fast and Slow focal views along with the Random view achieves the best top-1 accuracy of 80\%.
This outperforms configurations using only Fast views (78.6\%) or only Slow views (77.5\%) along with the Random view, highlighting the importance of learning from different temporal granularities.
Additional experimental results are provided in the appendix.
\section{Conclusion}
In this work, we introduce \ours, the first adaptation of Masked Siamese Networks to video representation learning.
By representing video frames as 2D super images composed of frames sampled from videos, the method enables standard 2D Vision Transformers to learn effective spatio-temporal representations without relying on computationally expensive 3D architectures.
The proposed decoder-free formulation replaces pixel-level reconstruction with feature alignment between masked and unmasked views, significantly reducing the computational cost of self-supervised pretraining.
Our experiments demonstrate that \ours~not only outperforms competing methods but does so with very less pre-training with videos.
\ours{} demonstrates strong performance in low-shot classification confirming the quality and transferability of the learned representations.
More broadly, our findings indicate that strong image foundation models can be efficiently adapted to the video domain through lightweight self-supervised learning, reducing the need for extensive video pretraining while maintaining competitive performance.
We hope this work motivates further research into efficient video adaptation strategies that leverage the rapidly growing ecosystem of image-pretrained foundation models.
\section{Acknowledgement}

This work was partially supported by ANRF Grant CRG/2023/005010. We acknowledge the National Supercomputing Mission (NSM) for providing computational resources via the DGX GPU Cluster at IIT Kharagpur and C-DAC for providing additional computing support through the ParamRudra cluster.

\bibliography{videomsn}

\newpage
\begin{appendices}
\section{Dataset Description}
\label{supp_sec:dataset}

\noindent\textbf{UCF101.} The UCF101~\cite{soomro2012ucf101} dataset is a widely-used benchmark for action recognition, comprising 13,320 unconstrained video clips collected from YouTube. It spans 101 human action categories, hierarchically grouped into five broad types: (i) human-object interaction, (ii) body-motion only, (iii) human-human interaction, (iv) playing musical instruments, and (v) sports activities. Each video is encoded at 25 frames per second (fps) with a resolution of 320$\times$240 pixels, and has an average duration of approximately 7.2 seconds. The dataset is structured into 25 groups, where each group contains 4 to 7 clips per action class, sharing commonalities in background, camera viewpoint, and other visual conditions. This grouping aids in benchmarking models under intra-class variations. The dataset is publicly accessible at: \href{https://www.crcv.ucf.edu/data/UCF101.php}{https://www.crcv.ucf.edu/data/UCF101.php}.

\noindent\textbf{HMDB51.} The HMDB51~\cite{hmdb} dataset (Human Motion Database) is a curated collection of 6,766 video clips, covering 51 distinct human action categories, with a minimum of 101 clips per class. Videos are sourced from a variety of real-world settings including movies, public databases, and YouTube. All clips are standardized to 30 fps, with the frame height fixed at 240 pixels, and width adjusted to preserve aspect ratio. The action categories are organized into five high-level groups: (i) General Facial Actions, (ii) Facial Actions with Object Manipulation, (iii) General Body Movements, (iv) Body Movements with Object Interaction, and (v) Body Movements for Human Interaction. This dataset poses significant challenges due to its diverse scenes, viewpoints, and motion dynamics, making it a valuable benchmark for action recognition research.
It is publicly available at: \href{https://serre-lab.clps.brown.edu/resource/hmdb-a-large-human-motion-database/}{https://serre-lab.clps.brown.edu/resource/hmdb-a-large-human-motion-database/}.

\noindent\textbf{Kinetics-400.} The Kinetics-400~\cite{kay2017kinetics} dataset, curated by DeepMind, is one of the largest and most diverse datasets for video action recognition. It comprises 306,245 video clips uniformly distributed across 400 action categories, with each class containing at least 400 clips. Videos have an average duration of 10 seconds, and depict actions in unconstrained environments sourced from YouTube. The actions are broadly categorized into three types: (i) person actions (e.g., drawing, drinking, laughing), (ii) person-person interactions (e.g., hugging, kissing, shaking hands), and (iii) person-object interactions (e.g., mowing the lawn, opening a present, washing dishes). Kinetics-400 sets a high standard for large-scale video understanding, particularly in real-world, diverse contexts. The dataset is publicly available at: \href{https://deepmind.com/research/open-source/kinetics}{https://deepmind.com/research/open-source/kinetics}.

\noindent\textbf{Something-Something V2.} The Something-Something V2 (SSV2)~\cite{goyal2017something} dataset is a challenging, large-scale benchmark for video action recognition. With over 220K videos and 174 action classes, it focuses on fine-grained human-object interactions. Unlike other datasets, SSV2 is considered motion-heavy due to its emphasis on the nuanced gestures and directional aspects of actions, such as "turning something upside down" or "pushing something from left to right". This design makes it a rigorous test for models that need to understand motion and context.  The dataset is publicly available at: \href{https://www.qualcomm.com/developer/software/something-something-v-2-dataset}{https://www.qualcomm.com/developer /software/something-something-v-2-dataset}.

\begin{table*}[t]
\centering
\captionsetup[table]{labelformat=empty}

\begin{minipage}[t]{0.45\linewidth}
\centering
\begin{tabular}{lc}
lr & Top-1\\
\shline
1e-4 & 68.6\\
1e-5 & \cellcolor{cyan!8}\textbf{69.1}\\
1e-6 & 66.2\\
\end{tabular}
\vspace{2mm}
\captionof{table}{\textbf{(a) Learning Rate:} We evaluate different learning rates and observe that $1e^{-5}$ yields the best performance.}
\addtocounter{table}{-1}
\sublabelabl{tab:effect_lr}{a}
\end{minipage}
\hfill
\begin{minipage}[t]{0.45\linewidth}
\centering
\begin{tabular}{lc}
Weight Decay & Top-1\\
\shline
0.05 & 68.8\\
0.01 & \cellcolor{cyan!8}\textbf{69.1}\\
0.1  & 68.9\\
\end{tabular}
\vspace{2mm}
\captionof{table}{\textbf{(b) Weight Decay:} A decay value of 0.01 provides the best trade-off.}
\addtocounter{table}{-1}
\sublabelabl{tab:effect_weight_decay}{b}
\end{minipage}
\medskip

\vspace{1mm}
\newsavebox{\focalviewbox}
\sbox{\focalviewbox}{%
    \begin{tabular}{ccc}
    \# Views & Throughput & Top-1\\
    \shline
    6 & 42 v/s & \cellcolor{cyan!8}\textbf{69.1}\\
    8 & 35 v/s & 69.2\\
    10 & 32 v/s & 69.3\\
    \end{tabular}%
}
\begin{minipage}[t]{0.45\linewidth}
\centering
\vbox to \ht\focalviewbox{%
    \vfill
    \begin{tabular}{ccc}
    Patch Drop & BS/GPU & Top-1\\
    \shline
    0.3 & 8  & 68.9\\
    0.5 & 10 & 68.9\\
    0.7 & 12 & \cellcolor{cyan!8}\textbf{69.1}\\
    0.9 & 14 & 68.8\\
    \end{tabular}
    \vfill
}
\vspace{15mm}
\captionof{table}{\textbf{(c) Patch Drop:} Increasing patch drop enables larger batch sizes and improves generalization.}
\addtocounter{table}{-1}
\sublabelabl{tab:effect_patch_drop}{c}
\end{minipage}
\hfill
\begin{minipage}[t]{0.45\linewidth}
\centering
\usebox{\focalviewbox}
\vspace{2mm}
\captionof{table}{\textbf{(d) Focal Views:} More views improve accuracy but reduce throughput. We select 6 views for the best trade-off.}
\addtocounter{table}{-1}
\sublabelabl{tab:effect_focal_view}{d}
\end{minipage}
\captionsetup[table]{labelformat=default}
\vspace{2mm}
\caption{Experiments to determine optimal hyperparameter settings on a class-wise uniformly sampled 25\% subset of the \textbf{Kinetics-400} dataset with \ours-DeiT (ViT-S) using $4\times 4$ superimage. We analyze the effect of (a) learning rate, (b) weight decay, (c) patch drop, and (d) focal views. The default configurations (highlighted) are the ones that achieves the best overall performance.}
\vspace{-6mm}
\label{tab:ablations}
\end{table*}

\section{Impact of Hyperparameters}
In this section, we analyze the impact of key model hyperparameters, including
(a) learning rate, (b) weight decay, (c) patch drop ratio, and (d) the number of focal views on Kinetics-400 dataset using \ours-DeiT with ViT-S backbone as the default backbone. Experiments, in the appendix are run using a single fixed seed due to computational constraints.

\subsection{Effect of Learning Rate (lr)} Table~\ref{tab:effect_lr} presents an ablation study evaluating the influence of different learning rates (lr) on the Top-1 accuracy (Top-1).
A learning rate of $1e-5$ yields the highest Top-1 accuracy of 69.1\%, indicating it is the most optimal among the tested configurations.
Increasing the learning rate to $1\mathrm{e}{-4}$ lowers accuracy to 68.6\%, while decreasing it to $1\mathrm{e}{-6}$ further drops accuracy to 66.2\%.

\subsection{Effect of Weight Decay} Table~\ref{tab:effect_weight_decay} reports the impact of varying weight decay values on Top-1 accuracy.
A weight decay of 0.01 achieves the best performance with 69.1\% Top-1 accuracy. Using 0.05 or 0.1 slightly reduces performance to 68.8\% and 68.9\%, respectively.
This suggests that 0.01 is an optimal choice for regularization in our setup.

\subsection{Effect of Patch Drop Ratio}
Table~\ref{tab:effect_patch_drop} presents an ablation study on the impact of varying patch drop ratios, where a fixed proportion of input spatio-temporal patches are masked (Tube masking) during training.
Increasing the patch drop ratio reduces the number of visible tokens passed to the encoder, thereby lowering per-sample memory and compute requirements.
This allows for larger batch sizes (BS) per GPU, as reflected in the table.
Among the tested values, a drop ratio of 0.7 yields the best Top-1 accuracy of 69.1\%, indicating an effective trade-off between information sparsity and model learning capacity (more towards increasing BS/GPU).
Lower ratios—0.3 and 0.5—retain 70\% and 50\% of the patches, respectively, and result in slightly reduced accuracies of 68.9\% for both, corresponding to performance drops of 0.2\% for both.
A higher ratio of 0.9 retains only 10\% of the patches, leading to a performance drop of 0.2\% (Top-1: 68.8\%) likely due to excessive loss of informative content.

\subsection{Effect of Number of Focal Views} Table~\ref{tab:effect_focal_view} presents an ablation study evaluating the impact of varying the number of focal views on both Top-1 accuracy and inference throughput (measured in videos per second).
As the number of focal views increases from 6 to 10, we observe a slight improvement in Top-1 accuracy—from 69.1\% to 69.3\%—suggesting that additional focal views provide marginal gains in performance.
However, this comes at the cost of significantly reduced throughput: from 42 videos/s at 6 views down to 32 videos/s at 10 views.
This trade-off highlights a key design consideration: while higher numbers of focal views can improve accuracy, they also impose greater computational overhead.
The configuration with 6 focal views offers the best balance between accuracy and efficiency, making it preferable in scenarios where both performance and scalability are critical.

\label{supp_sec:hyperparameter}

\section{Additional Ablations}
\label{supp_sec:additional_ablation}
Tables~\ref{tab:patch_drop},~\ref{tab:focal_view} presents a set of ablation experiments conducted on the full Kinetics-400 dataset using a ViT-S backbone.
The purpose of these studies is to validate and optimize key hyperparameters and design choices for the \ours-DeiT model.

\begin{table}[tb]
\centering

\resizebox{0.45\textwidth}{!}{
\begin{tabular}{ccc}
Patch Drop & Dataset & Top-1 Acc. (\%) \\
\hline
70\% & SSv2 & 64.8 \\

\cellcolor{cyan!8}50\% & \cellcolor{cyan!8} SSv2 & \cellcolor{cyan!8} 65.8 \\

25\% & SSv2 & 65.3 \\
\end{tabular}}
\vspace{2mm}
\caption{Effect of different \textbf{patch drop ratios} during the pretraining of our \ours-DINO model with a ViT-S backbone on the SSv2 dataset.}
\label{tab:patch_drop}
\end{table}
\begin{table}[tb]
\centering

\resizebox{0.55\textwidth}{!}{
\begin{tabular}{cc}
Focal Views & Top-1 Acc. (\%) \\
\hline
No Temporal Masking & 65.8 \\

Fast View Only & 65.7 \\

Slow View Only & 65.5 \\

\cellcolor{cyan!8} Both Views & \cellcolor{cyan!8} 65.8 \\
\end{tabular}}
\vspace{2mm}
\caption{Effect of different \textbf{focal view configurations} during pretraining of our \ours-DINO model with a ViT-S backbone using a patch drop ratio of 0.5 on the SSv2 dataset.}
\label{tab:focal_view}
\vspace{-10pt}
\end{table}
\begin{table}[tb]
\centering

\begin{tabular}{cc}
Masking Strategy & Top-1 Acc.(\%)\\
\shline
Random Masking & 79.0 \\
\cellcolor{cyan!8} Tube Masking   & \cellcolor{cyan!8} 80.0 \\
\noalign{\vskip\normalbaselineskip}
\end{tabular}
\vspace{2mm}
\captionof{table}{We compare the tube masking with the conventional random masking and observe a 1.0\% gain in top-1 accuracy. Tube masking enforces consistent spatiotemporal masking by applying the identical spatial masks across frames.}
\label{tab:tube_mask_supp}
\end{table}
\begin{table}[tb]
\centering

\begin{tabular}{cc}
No. of Prototypes & Top-1 \\
\shline
512  & 78.6 \\
1024 & \cellcolor{cyan!8}{80.0} \\
2048 & 80.0 \\
\end{tabular}
\vspace{2mm}
\captionof{table}{Effect of Number of Prototypes: On increasing the prototype count from 512 to 1024 provides a significant 1.4\% performance boost (78.6\% to 80.0\%). However, a further increase to 2048 yields no additional gain, establishing 1024 as the optimal configuration.}
\label{tab:proto_supp}
\end{table}

\subsection{Effect of Tube Masking.} 
We find that tube masking achieves better performance with an increment of 1\% (ref. Table~\ref{tab:tube_mask_supp}) than plain random masking. We attribute these interesting observations to the redundancy and temporal correlation in videos.
Tube masking effectively addresses the temporal redundancy in videos by masking continuous spatio-temporal regions, rather than random patches.
As seen in Fig. 2, from a super image $S^i$ when we get a Random View $R^i$, we decide for a random mask for the first frame and then apply the same mask in all the frames of the formed super image.
Then, all patches at the same spatial location across different temporal indices are either simultaneously masked or retained.
This reduces information leakage caused by frame-to-frame similarity and prevents the model from learning shortcut features.
As a result, it encourages the learning of meaningful spatio-temporal representations and enables successful training even with simple backbones like vanilla ViT on small-scale datasets.
These findings are consistent with the results of prior works like VideoMAE, which also demonstrated that tube masking significantly improves performance compared to randomly removing patches from all the frames in video masked image modeling.

\subsection{Effect of Number of Prototypes.} Table~\ref{tab:proto_supp} investigates the influence of the number of learnable prototypes of our \ours-DeiT (ViT-S) performance.
The purpose of this study is to validate the choice of this key hyperparameter and design choice for the \ours-DeiT model.
The results show that while increasing the number of prototypes from 512 to 1024 significantly improves the Top-1 accuracy from 78.6\% to 80.0\%, a further increase to 2048 does not yield any additional performance gain.
This confirms that 1024 prototypes is the optimal number for this model architecture, consistent with ~\cite{assran2022masked}.

\subsection{Ablation on Patch Drop Ratio}
Table~\ref{tab:patch_drop} analyzes the effect of different patch drop ratios during pretraining of our \ours-DINO (ViT-S) model.
A higher drop ratio of 70\% leads to a reduced performance of 64.8\%, indicating that excessive patch removal limits the model’s ability to learn meaningful representations.
Conversely, a smaller drop ratio of 25\% achieves 65.3\%, suggesting that insufficient patch dropping provides weaker regularization.
The best performance of \textbf{65.8\%} is obtained with a drop ratio of \textbf{50\%}, which provides an effective balance between representation learning and regularization.

\subsection{Ablation on Focal View Configurations}
Table~\ref{tab:focal_view} analyzes the impact of different focal view configurations during pretraining of our \ours-DINO (ViT-S).
Using only \textbf{Fast views} ($3\times3$ super-images) achieves a Top-1 accuracy of \textbf{65.7\%}, while using only \textbf{Slow views} ($2\times2$ super-images) yields \textbf{65.5\%}.
A configuration without temporal masking using six $4\times4$ focal views obtains \textbf{65.8\%} accuracy.
Our proposed setup, which \textbf{combines Fast and Slow views} (three $3\times3$ and three $2\times2$ super-image focal views), also achieves the \textbf{best performance of 65.8\%}, indicating that integrating multiple spatial scales improves representation learning.
\end{appendices}

\end{document}